\def\year{2020}\relax
\documentclass[letterpaper]{article} 
\usepackage{aaai20}  
\usepackage{times}  
\usepackage{helvet} 
\usepackage{courier}  
\usepackage[hyphens]{url}  
\usepackage{graphicx} 
\usepackage{graphicx}  
\usepackage{tikz}
\newcommand*\circled[1]{\tikz[baseline=(char.base)]{\node[shape=circle,draw,inner sep=1.5pt] (char) {#1};}}

\usepackage{amsmath}
\usepackage{amssymb}
\usepackage{attrib}
\usepackage{blkcntrl}
\usepackage{moredefs}
\usepackage{relsize}
\usepackage{booktabs}

\title{Multipurpose Intelligent Process Automation via Conversational Assistant}
\author{
  Alena Moiseeva\textsuperscript{1}, 
  Dietrich Trautmann\textsuperscript{1},
  Michael Heimann\textsuperscript{2},
  Hinrich Sch{\"u}tze\textsuperscript{1}\\
  \textsuperscript{1}Center for Information and Language Processing (CIS), 
  LMU Munich, Germany\\
  \textsuperscript{2}integral learning GmbH, Berlin, Germany\\
  {\tt alena@cis.lmu.de; inquiries@cislmu.org}\\}

\long\def\eat#1{}

\def\figref#1{Figure~\ref{fig:#1}}
\def\figlabel#1{\label{fig:#1}\label{p:#1}}

\def\tabref#1{Table~\ref{tab:#1}}
\def\tablabel#1{\label{tab:#1}\label{p:#1}}
\def\secref#1{Section \ref{sec:#1}}
\def\seclabel#1{\label{sec:#1}}
\def\eqref#1{Eq.~\ref{eqn:#1}}

\usepackage{booktabs}
\usepackage{multirow}
\usepackage{color}
\usepackage{hhline}

\begin{document}


\maketitle

\begin{abstract}
Intelligent Process Automation (IPA) is an emerging
technology with a primary goal to assist the knowledge
worker by taking care of repetitive, routine and
low-cognitive tasks. Conversational agents that can interact
with users in a natural language are a potential application for IPA systems. Such intelligent agents can assist the user by answering specific questions and executing routine tasks that are ordinarily performed in a natural language (i.e., customer support). 

In this work, we tackle a challenge of implementing an IPA conversational assistant in a real-world industrial setting with a lack of structured training data. Our proposed system brings two significant benefits: First, it reduces repetitive and time-consuming activities and, therefore, allows workers to focus on more intelligent processes. Second, by interacting with users, it augments the resources with structured and to some extent labeled training data. We showcase the usage of the latter by re-implementing several components of our system with Transfer Learning (TL) methods.
\end{abstract}


\section{Introduction}
\seclabel{intro}

Robotic Process Automation (RPA) is a type of software bots that simulates hand-operated human activities like entering data into a system, registering into accounts, and accomplishing straightforward but repetitive workflows \cite{a2018compete}. However, one of the drawbacks of RPA-bots is their susceptibility to changes in defined scenarios: being designed for a particular task, the RPA-bot is usually not adaptable to other domains or even light modifications in a workflow \cite{a2018compete}. This inability to readjust to shifting conditions gave rise to Intelligent Process Automation (IPA) systems. IPA-bots combine RPA with Artificial Intelligence (AI) and thus are able to execute more cognitively demanding tasks that require i.a. reasoning and language understanding.  Hence, IPA-bots advanced beyond automating shallow ``click tasks''  and can perform jobs more intelligently -- by means of machine learning algorithms. Such IPA-systems undertake time-consuming and routine tasks, and thus enable smart workflows and free up skilled workers to accomplish higher-value activities.

One of the potential applications of Natural Language
Processing (NLP) within the IPA domain are conversational
interfaces that enable human-to-machine interaction. The
main benefit of conversational systems is their ability to
give attention to several users simultaneously while
supporting natural communication. A conventional dialogue
system comprises multiple stages and involves different
types of NLP subtasks, starting with Natural Language
Understanding (NLU) (e.g., intent classification, named
entity extraction) and going towards dialogue management
(i.e., determining the next possible bot action, considering
the dialogue history) and response generation (e.g.,
converting the semantic representation of the next system action into a natural language utterance). A typical dialogue system for IPA purposes undertakes shallow customer support requests (e.g., answering of FAQs), allowing human workers to focus on more sophisticated inquiries.

Recent research in the dialogue generation domain is conducted by employing AI-techniques like machine and deep learning \cite{wen2016network,lowe2017training}. However, conventional supervised methods have limitations when applied to real-world data and industrial tasks. The primary challenge here refers to a training phase since a robust model requires an extensive amount of structured and labeled data, that is often not available for domain-specific problems. Especially if it concerns dialogue data, which has to be appropriately structured as well as labeled and annotated with additional information. Therefore, despite the popularity of deep learning end-to-end models, one still needs to rely on conventional pipelines in practical dialogue engineering, especially while setting a new domain. However, with few structured data available, transfer learning methods can be used. Such algorithms enable training of the systems with less or even a minimal amount of data, and are able to transfer the knowledge obtained during the training on existing data to the unseen domain.  

\subsection{Outline and Contributions} This paper addresses the challenge of implementing a dialogue system for IPA purposes within the practical e-learning domain with the initial absence of training data. 
Our contributions within this work are as follows:
\begin{itemize}
\itemsep0em 
\item We implemented a robust dialogue system for IPA purposes within the practical e-learning domain and within the conditions of \emph{missing training (dialogue) data} (see \secref{model} -- \secref{eval}). The system is currently deployed at the e-learning platform. 
\item The system has two purposes: 
\begin{itemize}
\itemsep0em 
    \item First, it reduces repetitive and time-consuming
    activities and, therefore, allows workers of the
    e-learning platform to focus solely on complex questions;
    \item Second, by interacting with users, it augments the resources with structured and to some extent labeled training data for further possible implementation of learnable dialogue components (see \secref{structured});
\end{itemize}
\item We showcased that even a small amount of structured dialogues could be successfully used for re-training of dialogue units by means of Transfer Learning techniques (see \secref{deep}).
\end{itemize}

\section{Target Domain \& Task Definition}
\label{ch:omb}

\emph{OMB+}\footnote{This work was conducted in collaboration with \emph{OMB+}.} is a German e-learning platform that assists students who are preparing for an engineering or computer science study at a university.  The central purpose of the course is to support students in reviving their mathematical skills so that they can follow the upcoming university courses. The platform is thematically segmented into $13$ sections and includes free mathematical classes with theoretical and practical content. Besides that, \emph{OMB+}\footnote{https://www.ombplus.de/} provides a possibility to get assistance from a human tutor via a \emph{chat interface}.  Usually, the students and tutors interact in written form, and the language of communication is German. The current problem of the \emph{OMB+} platform is that the number of students grows every year, but to hire more qualified human tutors is challenging and expensive. This results in a more extended waiting period for students until their problems can be considered. 

In general,  student questions can be grouped into three main categories: \emph{organizational questions} (e.g., course certificate), \emph{contextual questions} (e.g., content, theorem) and \emph{mathematical questions} (e.g., exercises, solutions). To assist a student with a mathematical question, a tutor has to know the following regular information: What kind of \emph{topic} (or \emph{sub-topic}) a student has a problem with. At which \emph{examination mode} (i.e., quiz, chapter level training or exercise, section level training or exercise, or final examination) the student is working right now.  And finally, the \emph{exact question number} and \emph{exact problem formulation}. This means that a tutor has to request the same information every time a new dialogue opens, which is very time consuming and could be successfully solved by means of an IPA dialogue bot.

\section{Model}
\seclabel{model}
The main objective of the proposed system is to interact with students at the beginning of every conversation and gather information on the topic (and sub-topic), examination mode and level, question number and exact problem formulation.  Therefore, the system saves time for tutors and allows them to handle solely complex mathematical questions. Besides that, the system is implemented in a way such that it accumulates labeled dialogues in the background and stores them in a structured form.

\subsection{Dialogue Modules}
\seclabel{modules}
\figref{fig:df} (see the Appendix) displays the entire
dialogue flow. In a nutshell, the system receives a user
input, analyzes it and extracts information, if provided. If
some of the required information is missing, the system asks the student to provide it. When all the information is collected, it will be automatically validated and subsequently forwarded to a human tutor, who then can directly proceed with the assistance. In the following we will describe the central components of the system.
\vspace{0.5mm}

{\bf OMB+ Design:} \figref{fig:omb} (see \secref{ombfig} of the Appendix) illustrates the internal structure and design of the \emph{OMB+} platform. It has \emph{topics} and \emph{sub-topics}, as well as four \emph{examination modes}. Each \emph{topic} (\figref{fig:omb}, tag $1$) corresponds to a chapter level and always has \emph{sub-topics} (\figref{fig:omb}, tag $2$), which correspond to a section level.  Examination modes \emph{training} and \emph{exercise} are ambiguous, because they correspond to either a \emph{chapter} (\figref{fig:omb}, tag $3$) or a \emph{section} (\figref{fig:omb}, tag $5$) level, and it is important to differentiate between them, since they contain different types of content. The mode \emph{final examination} (\figref{fig:omb}, tag $4$) always corresponds to a chapter level, whereas \emph{quiz} (\figref{fig:omb}, tag $5$) can belong only to a section level. According to the design of the \emph{OMB+} platform, there are several ways of how a possible dialogue flow can proceed. 
\vspace{0.5mm}

{\bf Preprocessing:} In a natural language dialogue, a user may respond in many different ways, thus, the extraction of any data from user-generated text is a challenging task due to a number of misspellings or confusable spellings (e.g., \emph{Exercise \underline{1.a}}, \emph{Exercise \underline{1 (a)}}). Therefore, to enable a reliable extraction of entities, we preprocessed and normalized (e.g., misspellings, synonyms) every user input before it was sent to the Natural Language Understanding (NLU) module.
The preprocessing includes following steps:
\begin{itemize}
\itemsep0em 
\item lowercasing and stemming of all words in the input;
\item removal of German stop words and punctuation;
\item all mentions of $x$ in mathematical formulas were removed to avoid confusion with roman number 10 (``$X$'');
\item in a combination of the type: word \emph{``Chapter/Exercise''} + digit written as a word (i.e \emph{``first''}, \emph{``second''}), word was replaced with a digit (\emph{``in first Chapter''} $\rightarrow$ \emph{``in Chapter 1''}), roman numbers were replaced with digits as well (\emph{``Chapter IV''} $\rightarrow$ \emph{``Chapter 4''}).
\item detected ambiguities were normalized (e.g., \emph{``Trainingsaufgabe''} $\rightarrow$ \emph{``Training''})\footnote{Translates to ``training exercise''};
\item recognized misspellings resp.~typos were corrected (e.g., \emph{``Di\underline{feernzial}rechnung''}
$\rightarrow$ \emph{``Di\underline{fferential}rechnung'})\footnote{Translates to ``differential calculus''}' 
\item permalinks were parsed and analyzed. From each permalink it is possible to extract topic, examination mode and question number;
\end{itemize} 
\vspace{0.5mm}

{\bf Natural Language Understanding (NLU):} We implemented an NLU unit utilizing handcrafted rules, Regular Expressions (RegEx) and Elasticsearch\footnote{\url{https://www.elastic.co/products/elasticsearch}} (ES) API. The NLU module contains following functionalities: 

\begin{itemize}
\itemsep0em 
\item  {\bf Intent classification:} As we mentioned above, student questions can be grouped into three main categories: Organizational questions, contextual questions and mathematical questions. To classify the input message by its category or so-called \emph{intent}, we utilized key-word information predefined by handcrafted rules. We assumed that particular words are explicit and associated with a corresponding intent. If no intent could be classified, then it is assumed that the NLU unit was not capable of understanding and the intent is interpreted as unknown. In this case, the system requests the user to provide an intent manually (by picking one from the mentioned three options). The questions from organizational and theoretical categories are directly delivered to a human tutor, while mathematical questions are processed by the automated system for further analysis.
\item {\bf Entity Extraction:} Next, the system attempts to
retrieve the entities from a user message on the topic (and
sub-topic), examination mode and level, and question
number. This part is implemented using Elasticsearch (ES) and RegEx. To enable the use of ES, we indexed the \emph{OMB+} site to an internal database. Besides indexing the topics and titles, we also provided information on possible synonyms or writing styles. We additionally filed \emph{OMB+} permalinks, which direct to the site pages. To query the resulting database, we utilized the internal Elasticsearch \emph{multi\underline{ }match} function and set the \emph{minimum\underline{ }should\underline{ }match} parameter to $20\%$. This parameter defines the number of terms that must match for a document to be considered relevant. Besides that, we adopted \emph{fuzziness} with the maximum edit distance set to $2$ characters. The fuzzy query uses similarity based on Levenshtein edit distance \cite{levenshtein1966binary}. Finally, the system generates a ranked list of possible matching entries found in the database within the predefined \emph{relevance\underline{ }threshold} (we set it to $\theta$=$1.5$). We pick the most probable entry as the correct one and extract the corresponding entity from the user input.
\end{itemize}
To summarize, the NLU module receives the user input as a preprocessed text and checks it across all predefined RegEx statements and for a match in the Elasticsearch database. Every time the entity is extracted, it is entered in the \emph{Information Dictionary} (ID). The ID has the following six slots to be filled in: topic, sub-topic, examination level, examination mode, question number, and exact problem formulation.
\vspace{0.5mm}

{\bf Dialogue Manager} consists of the Dialogue State
Tracker (DST), that maintains a representation of the
current dialog state, and of the Policy Learner (PL) that
defines the next system action. In our model, the
system's \emph{next action} is defined by the state of the
previously obtained information stored in the Information
Dictionary. For instance, if the system recognizes that the
student works on the final examination, it also understands
(defined by the logic in the predefined rules) that there is
no need to ask for sub-topic because the final examination
always corresponds to a chapter level (due to the design
of \emph{OMB+} platform). If the system identifies that the
user has difficulties in solving a quiz, it has to ask for
the corresponding topic and sub-topic if not yet provided by
a user (because the quiz always refers to a section level). To determine all of the potential dialogue flows, we implemented \emph{Mutually Exclusive Rules} (MER), which indicate that two events $e_{1}$ and $e_{2}$ are mutually exclusive or disjoint if they cannot both occur at the same time (thus, the intersection of these events is empty: $P(A \cap B) = 0$). Additionally, we defined \emph{transition} and \emph{mapping rules}. The formal explanation of rules can be found in \secref{mers} of the Appendix. Following the rules, we generated $56$ \emph{state transitions}, which define next system actions. Being on a new dialogue state, the system compares the extracted (i.e., updated) information in the ID with the valid dialogue states (see \secref{mers} of the Appendix for the explanation of the validness) and picks the mapped action as the next system's action.

The abovementioned rules are intended to support the current design of the \emph{OMB+} learning platform. However, additional MERs could be added to generate new transitions. Exemplifying this, we conducted experiments with the former design of the platform, where all the topics, except for the first one, had only \emph{sub-topics}, whereas the first topic had both \emph{sub-topics} and \emph{sub-sub-topics}. We could  effortlessly generate the missing transitions with our approach. The number of possible transitions, in this case, increased from $56$ to $117$.
\vspace{0.5mm}

{\bf Meta Policy:} Since the system is intended to operate in a real-world scenario, we had to implement additional policies that control the dialogue flow and validate the system's accuracy. Below we describe these policies:
\begin{itemize}
\itemsep0em 
\item {\bf Completeness of the Information Dictionary:} In
this step, the system validates the completeness of the ID,
which is defined by the number of obligatory slots filled in
the information dictionary.  There are $6$ distinct cases
when the ID is considered to be complete
(see \secref{ap:a_cases} of the Appendix). For instance, if
a user works on a final examination, the system does not has
to request a sub-topic or examination level.  Thus, the ID
has to be filled only with data for a topic, examination
mode, and question number, whereas, if the user works on a quiz, the system has to gather information about the topic, sub-topic, examination mode, and the question number. Once the ID is complete, it is provided to the  \emph{verification step}. Otherwise, the system proceeds according to the next action. The system extracts the information in each dialogue step, and thus if the user provides updated information on any subject later in the dialogue, the corresponding slot will be \emph{updated} in the ID.

\item {\bf Verification Step:} Once the system has obtained all the necessary information (i.e., ID is complete), it proceeds to the final verification. In that step, the collected data is shown to the student in the session.  The student is asked to verify the correctness of the collected data and if some entries are wrong, to correct them. The ID is, where necessary, updated with the user-provided data.  This procedure repeats until the user confirms the correctness of the assembled data.

\item {\bf Fallback Policy:} In some cases, the system fails to derive information from a student query, even if the student provides it.  It is due to the Elasticsearch functionality and previously unseen RegEx patterns.  In these cases, the system re-asks a user and attempts to retrieve information from a follow-up query.  The maximum number of re-ask attempts is set to three times ($r=3$). If the system is unable to extract information after three times, the user input is considered as the \emph{ground truth} and saved to the appropriate slot in ID.  An exception to this rule applies where the user has to specify the intent manually. In this case, after three unclassified attempts, a session is directly handed over to a human tutor.

\item {\bf Human Request:} In each dialogue state, a user can switch to a human tutor.  For this, a user can enter the \emph{human} key-word.  Hence, every user message is additionally analyzed for the presence of this key-word.
\end{itemize}
\vspace{0.5mm}

{\bf Response Generation:} In this module, the semantic representation of the system's next action is transformed into natural language. Hence, each possible action is mapped to precisely one utterance, which is stored in the templates. Some of the predefined responses are fixed (i.e., \emph{``Welches Kapitel bearbeitest du gerade?'')\footnote{Translates to: ``Which chapter are you currently working on?''}}, others have \emph{placeholders} for system values. In the latter case, the utterance can be formulated dependent on the actual ID. The dialogue showcases can be found in \secref{ap:a_appendix} of the Appendix.

\section{Evaluation}
\seclabel{eval}
In order to get the feedback on the quality, functionality,
and usefulness of the introduced model, we evaluated it in
two ways: first, with an automated method using $130$
manually annotated dialogues, to prove the robustness of the
system,  and second -- with human tutors from \emph{OMB+} -- to investigate the user experience. We describe the details as well as the most common errors below.

\subsection{Automated Evaluation}
\seclabel{ch:a_eval_rb}
To conduct an automated evaluation, we manually created a dataset with $130$ dialog scenarios. We took real first user questions and predefined possible user responses, as well as gold system answers. These scenarios cover most frequent dialogues that we previously saw in human-to-human dialogue data. We tested our system by implementing a \emph{self-chatting evaluation bot}. The evaluation cycle could be described as follows: 
\begin{itemize}
\itemsep0em 
\item The system receives the first question from a predefined dialogue via an API request, preprocesses and analyzes it in order to extract entities.
\item Then it estimates the applicable next action, and responds according to it.
\item This response is then compared to the system's gold answer: If the predicted answer is correct, then the system receives the next predefined user input from the dialog and responds again, as defined above. This procedure continues until the dialog terminates (i.e., ID is complete). Otherwise, the system fails, reporting the unsuccessful case number.
\end{itemize}  

Our final system successfully passed this evaluation for all $130$ cases.

\subsection{Human Evaluation and Error Analysis}
\seclabel{h_eval_rb}
To evaluate our system on irregular examples, we conducted experiments with human tutors from the \emph{OMB+} platform. The tutors are experienced regarding rare or complex questions, ambiguous answers, misspellings and other infrequent but still very relevant problems, which occur during a dialogue. In the following, we investigate some common errors and make additional observations.
\vspace{0.5mm}

{\bf Misspellings and confusable spellings} occur quite often in the user-generated text, and since we attempt to let the conversation remain very natural from the user side and thus, cannot require formal writing, we have to deal with various writing issues. One of the most frequent problems is misspellings. German words are generally long and can be complicated, and since users type quickly, this often leads to the wrong order of characters within words. To tackle this challenge, we used \emph{fuzzy\underline{ }match} within ES. However, the maximum allowed edit distance in \emph{Elasticsearch} is set to $2$ characters.  This means, that all the misspellings beyond this threshold could not be accurately recognized by ES (e.g., \emph{Differ\underline{ent}ialrechnung} vs
\emph{Differ\underline{nez}ialrechnung}). Another characteristic example would be the writing of the section or question number. The equivalent information can be written in several distinct ways, which has to be considered in our RegEx unit (e.g., \emph{Exercise 5 a}, \emph{Exercise V a}, \emph{Exercise 5 (a)}).  A similar problem occurs with confusable spelling (i.e.: \emph{Differential\underline{rechnung}} vs
\emph{Differential\underline{gleichung}}). We analyzed the cases mentioned above and added some of the most common issues to the ES database or handled them with RegEx during the preprocessing step.
\vspace{0.5mm}

{\bf Elasticsearch Threshold:} 
In some cases, the system failed to extract information, although the user provided it.  In other cases, ES extracts information that was not mentioned in a user query at all.  That occurs due to the \emph{relevancy\underline{ }scoring} algorithm of Elasticsearch, where a document's score is a combination of textual similarity and other metadata based scores. Our analysis revealed that ES mostly fails to extract the information if the sentence (i.e., user message) is quite short (e.g., $5$ words). To overcome this difficulty, we combined the current input $u_{t}$ with the dialog history. This step eliminated the problem and improved the retrieval quality. To solve the case where Elasticsearch extracts incorrect information (or information that was not mentioned in a query) is more challenging. We discovered that the problem comes from short words or sub-words (e.g., suffixes, prefixes), which ES considers to be credible enough. The Elasticsearch documentation suggests getting rid of stop words to eliminate this behavior.  However, this did not improve the search in our case.  Also, fine-tuning of ES parameters such as the \emph{relevance\underline{ }threshold}, \emph{prefix\underline{ }length\footnote{The number of initial characters which will not be fuzzified. It helps to reduce the number of terms which must be examined.}} and \emph{minimum\underline{ }should\underline{ }match\footnote{Indicates a number of terms that must match for a document to be considered relevant.}} parameter did not bring significant improvements. To cope with this problem, we implemented a \emph{verification step}, where a user is given a chance to correct the erroneously retrieved information.

The overall feedback from the tutors  included reduced repetitive activities as well as reduced waiting times for students until their questions were processed. Also, tutors reported that the rate of cancelled sessions (switching to a human tutor) is rather low.

\section{Structured Dialogue Acquisition}
\seclabel{structured}
As we already mentioned, our system attempts to support the human tutor by assisting students, but it also collects structured and labeled training data in the background. In a \emph{trial run} of the rule-based system, we were able to accumulate a \emph{toy-dataset} with training dialogues. The assembled dialogues have the following format:
\begin{itemize}
\itemsep0em 
\item Plain dialogues with unique dialogue indexes;
\item Plain Information Dictionary information (e.g., extracted entities) collected for the whole dialogue; 
\item Pairs of questions (i.e.,  user requests) and responses (i.e., bot responses) with the unique dialogue- and turn-indexes;
\item Triples in the form of \emph{(User Request, Next Action, Response)}. Information on the next system's action could be employed to train a Dialogue Manager unit with (deep-) machine learning algorithms;
\item For each state in the dialogue, we saved the entities that the system was able to extract from the provided user query, along with their position in the utterance. This information could be used to train a custom, domain specific Named Entity Recognition model.
\end{itemize}

\section{Re-implementation of units with BERT}
\seclabel{deep}
As we mentioned before, there are many cases, especially in the industry, where the labeled and structured data is not directly available. Collecting and labeling such data is often a tedious and time-consuming task. Thus, algorithms that enable training of the systems with less or even a minimal amount of data are highly required. Such algorithms can transfer the knowledge obtained during the training on existing data to the unseen domain. They are, therefore, one of the potential solutions for industrial problems. 

Once we assembled a dataset of structured data via our rule-based system, we re-implemented two out of three central dialogue components in our conversational assistant with deep learning methods.  Since the available data was collected in a trial-run and thus the obtained dataset was rather small to train a machine learning model from scratch, we utilized the Transfer Learning approach, and fine-tuned the existing pre-trained model (i.e., BERT) for our target domain and data.

For the experiments, we defined two tasks:
\begin{itemize}
\itemsep0em 
\item First, we studied the \emph{Named Entity Recognition}
problem in a custom domain setting. We defined a sequence
labeling task and employed the BERT model \cite{devlin2018bert}. We applied the model to our dataset and fine-tuned it for \emph{six} ($6$) domain-specific (i.e., e-learning) entities and \emph{one} ($1$) ``unknown'' label. 
\item Second, we investigated the effectiveness of BERT for the dialogue manager core. For that experiment, we defined a classification task and applied the model to predict the system's \emph{Next Action} for every given user utterance in a conversation. We then computed the macro F-score for $13$ possible actions and an average dialogue accuracy. 
\end{itemize}

Finally, we verified that the applied model performed well on both tasks: We achieved the performance of $0.93$ macro F1 points for Named Entity Recognition (NER) and $0.75$ macro F1 points for the Next Action Prediction (NAP) task. We, therefore, conclude that both NER and NAP components could be employed to substitute or extend the existing rule-based modules. 
\vspace{0.5mm}

{\bf Data \& Descriptive Statistics:}
The dataset that we collected during the trial-run consists of $300$ structured dialogues with the average length of a dialogue being \emph{six} ($6$) utterances. Communication with students was performed in the German language. Detailed general statistics can be found in \tabref{statistics}. 

\begin{table}[h]
\centering
\begin{tabular}{@{}lr@{}}
\toprule
 & \multicolumn{1}{r}{\textbf{Value}} \\ \midrule
Max. Len. Dialogue (in utterances) & $15$ \\
Avg. Len. Dialogue (in utterances) & $6$ \\
Max. Len. Utterance (in tokens) & $100$ \\
Avg. Len. Utterance (in tokens) & $9$ \\
\# Overall Unique Action Labels & $13$ \\
\# Overall Unique Entity Labels & $7$ \\ \midrule 
{\bf Train} -- \# Dialogues (\# Utterances) & $200$ ($1161$) \\
{\bf Eval}\phantom{i} -- \# Dialogues (\# Utterances) & $50$ ($279$) \\
{\bf Test}\phantom{i.} -- \# Dialogues (\# Utterances) & $50$ ($300$)
\end{tabular}
\caption{General statistics for conversational dataset.}
\tablabel{statistics}
\end{table}
\vspace{0.3cm}

\begin{table}[h]
\centering
\begin{tabular}{@{}lrlr@{}}
\toprule
\textbf{Action} & \textbf{Count} & \textbf{Action} & \textbf{Count} \\ \midrule
Final Request & \multicolumn{1}{r|}{321} & Unk. & \multicolumn{1}{r}{80} \\
Human Handover & \multicolumn{1}{r|}{300} & Subtopic & \multicolumn{1}{r}{55} \\
Exact Question & \multicolumn{1}{r|}{286} & Correct Request & \multicolumn{1}{r}{40} \\
Question Number & \multicolumn{1}{r|}{176} & Verify Request & \multicolumn{1}{r}{34} \\
Examination & \multicolumn{1}{r|}{175} & Org. & \multicolumn{1}{r}{17} \\
Topic & \multicolumn{1}{r|}{137} & Text. & \multicolumn{1}{r}{13} \\
Level & \multicolumn{1}{r|}{130} &  &  \\ \bottomrule
\end{tabular}
\caption{Detailed statistics on possible systems actions. Column ``Count'' denotes the number of occurrences of each action in the entire dataset.}
\tablabel{astatistics}
\end{table}

\begin{table}[h]
\centering
\begin{tabular}{@{}lr@{}}
\toprule
\textbf{Entity}\phantom{xxxx} & \textbf{Count} \\ \midrule
Question Nr. \phantom{xxxxxxx} &  \multicolumn{1}{r}{317} \\
Chapter & \multicolumn{1}{r}{311} \\
Examination & \multicolumn{1}{r}{303} \\
Subtopic & \multicolumn{1}{r}{198} \\
Level & \multicolumn{1}{r}{80} \\
Intent & \multicolumn{1}{r}{70} \\ \bottomrule
\end{tabular}
\caption{Detailed statistics on possible named entities. Column ``Count'' denotes the number of occurrences of each entity in the entire dataset.}
\tablabel{estatistics}
\end{table}
\vspace{0.5mm}

{\bf Named Entity Recognition:}  
We defined a sequence labeling task to extract custom
entities from user input. We assumed seven ($7$) possible
entities (see \tabref{estatistics}) to be recognized by the
model: \emph{topic}, \emph{subtopic}, \emph{examination
mode} and \emph{level}, \emph{question
number}, \emph{intent}, as well as the entity \emph{other}
for remaining words in the utterance. Since the data
obtained from the rule-based system already contains
information on the entities extracted from each user query
(i.e., by means of Elasticsearch), we could use it to train
a domain-specific NER unit. However, since the user-input
was informal, the same information could be provided in
different writing styles. That means that a single entity
could have different surface forms (e.g., synonyms, writing
styles) (although entities that we extracted from the
rule-based system were all converted to a universal
standard, e.g., official chapter names). To consider all of the variable entity forms while post-labeling the original dataset, we defined generic entity names (e.g., chapter, question nr.) and mapped variations of entities from the user input (e.g., Chapter = [Elementary Calculus, Chapter $I$, ...]) to them. \vspace{0.5mm} 

{\bf Next Action Prediction:} We defined a classification problem to predict the system's next action according to the given user input. We assumed $13$ custom actions (see \tabref{astatistics}) that we considered being our labels. In the conversational dataset, each input was automatically labeled by the rule-based system with the corresponding next action and the dialogue-id. Thus, no additional post-labeling was required. We investigated two settings:
\begin{itemize}
\item {\bf Default Setting:} Using only a user input and the corresponding label (i.e., next action) without additional context. By default, we run all of our experiments in this setting.
\item {\bf Extended Setting:} Using a user input, a corresponding next action, and a \emph{previous system action} as a source of additional context. For this setting, we run an experiment with the best performing model from the default setting.
\end{itemize} 

The overall dataset consists of $300$ labeled dialogues, where $200$ (with $1161$ utterances) of them were employed for training, and $100$ for evaluation and test sets ($50$ dialogues with about $300$ utterances for each set respectively). 
\vspace{0.5mm}

{\bf Model Settings:} 
For the NER task we conducted experiments with German and multilingual BERT implementations\footnote{https://github.com/huggingface/transformers}. Since in the German language the capitalization of words plays a significant role, we run our tests on the capitalized input, while keeping the original punctuation. Hence, we employed the available {\tt base model} for both multilingual and German BERT implementations in the {\tt cased} version. We set the learning rate for both models to $1e-4$ and the maximum length of the tokenized input was set to $128$ tokens. We run the experiments multiple times with different seeds for a maximum of $50$ epochs with the training batch size set to $32$. We utilized {\tt AdamW} as the optimizer and employed early stopping, if the performance did not change significantly after $5$ epochs.

For the NAP task we conducted experiments with German and multilingual BERT implementations as well. Here, we investigated the performance of both capitalized and lowercased input, as well as plain and preprocessed data. For the multilingual BERT, we employed the {\tt base model} in both {\tt cased} and {\tt uncased} variations. For the German BERT, we utilized the {\tt base model} in the {\tt cased} variation only\footnote{Uncased pre-trained variation of the model was not available.}. For both models, we set the learning rate to $4e-5$, and the maximum length of the tokenized input was set to $128$ tokens. We run the experiments multiple times with different seeds for a maximum of $300$ epochs with the training batch size set to $32$. We utilized {\tt AdamW} as the optimizer and employed early stopping, if the performance did not change significantly after $15$ epochs.
\vspace{0.5mm}

{\bf Evaluation and Discussion:} For the evaluation, we computed \emph{word-level} macro F1 score for the NER task and \emph{utterance-level} macro F1 score for the NAP task. The word-level F1 is estimated as the average of the F1 scores per class, each computed from all words in the evaluation and test sets. The results for the NER task are depicted in \tabref{resner}. 
For utterance-level F1, a single label (i.e., next action) is obtained for the whole utterance. The results for the NAP task are presented in \tabref{resnap}. 
We additionally computed \emph{average dialogue accuracy} for the best performing NAP models. This score denotes how well the predicted next actions match the gold next actions and thus form the dialogue flow within each conversation. The average dialogue accuracy was computed for $50$ dialogues in the evaluation and test sets respectively. The results are displayed in \tabref{avgdia}.

The obtained results for the NER task revealed that German BERT performed significantly better than the multilingual BERT model. The performance of the custom NER unit is at $0.93$ macro F1 points for all possible named entities (see \tabref{resner}). In contrast, for the NAP task, the multilingual BERT model obtained better performance than the German BERT model. Here, the best performing system in the default setting achieved a macro F1 of $0.677$ points for $14$ possible labels, whereas the model in the extended setting performed better -- its highest macro F1 score is $0.752$ for the same amount of labels (see \tabref{resnap}). 
Considering the dialogue accuracy, the extended system trained with multilingual BERT achieved better results than the default one with $0.801$ accuracy points compared to $0.724$ accuracy points for the test set (see \tabref{avgdia}).
The overall observation for the NAP is that the capitalized setting improved the performance of the model, whereas the inclusion of punctuation has not positively influenced the results.

\begin{table}[h!]
\centering
\begin{tabular}{@{}cccccccc@{}}
\toprule
\multicolumn{1}{c}{\textbf{Task}} & \multicolumn{1}{c}{\textbf{Model}} & \multicolumn{1}{c}{\textbf{Cased}} & \multicolumn{1}{c}{\textbf{Punct.}} & \multicolumn{1}{c}{\textbf{Ext.}} & \multicolumn{2}{c}{\textbf{F1}} \\ \midrule
& & & & & {\bf Eval} & {\bf Test}   \\ \midrule
NAP & GER & $\checkmark$ & $\times$ & $\times$ & 0.711 &  0.673\\
NAP & GER & $\checkmark$ & $\checkmark$ & $\times$ & 0.701 & 0.606\\
NAP & Mult & $\checkmark$ & $\checkmark$ & $\times$ & 0.688 & 0.625\\
NAP & Mult & $\checkmark$ & $\times$ & $\times$ & \underline{0.769} & \underline{0.677}  \\
NAP & Mult & $\checkmark$ & $\times$ & $\checkmark$ & \bf{0.810} & \bf{0.752}  \\
NAP & Mult & $\times$ & $\checkmark$ & $\times$ & 0.664 & 0.596\\
NAP & Mult & $\times$ & $\times$ &  $\times$ & 0.742 & 0.502\\
\bottomrule
\end{tabular}
\caption{Utterance-level F1 for the NAP task. Underlined: best performance for evaluation and test sets for default setting (without previous action context). In bold: best performance for evaluation and test sets on extended setting (with previous action context).}
\tablabel{resnap}
\end{table}

\begin{table}[h!]
\centering
\begin{tabular}{@{}ccccccc@{}}
\toprule
\multicolumn{1}{c}{\textbf{Task}} & \multicolumn{1}{c}{\textbf{Model}} & \multicolumn{1}{c}{\textbf{Cased}} & \multicolumn{1}{c}{\textbf{Punct.}} &  \multicolumn{2}{c}{\textbf{F1}} \\ \midrule
& & & & {\bf Eval} & {\bf Test}  \\ \midrule
NER & GER & $\checkmark$ & $\checkmark$ & \bf{0.971} & \bf{0.930} \\
NER & Mult & $\checkmark$ & $\checkmark$ & 0.926 & 0.905 \\  \bottomrule
\end{tabular}
\caption{Word-level F1 for the NER task. In bold: best performance for evaluation and test sets.}
\tablabel{resner}
\end{table}

\begin{table}[h]
\centering
\begin{tabular}{@{}lll@{}}
\toprule
{\bf Model} & \multicolumn{2}{c}{{\bf Accuracy}} \\ \midrule
 & {\bf Eval} & \multicolumn{1}{l}{\bf Test} \\ \midrule
{\bf NAP default \phantom{xxxxx}} & 0.765 & 0.724  \\
{\bf NAP extended \phantom{xxxx}} & \bf{0.813} & \bf{0.801} \\ \bottomrule
\end{tabular}
\caption{Average dialogue accuracy computed for the NAP task for best performing models. In bold: best performance for evaluation and test sets.}
\tablabel{avgdia}
\end{table}

\section{Error Analysis:} After the evaluation step, we analyzed the cases, where the model failed to predict the correct action or labeled the named entity span erroneously. Below we describe the most common errors for both tasks.

\textbf{Next Action Prediction:} One of the most frequent errors in the \emph{default model} was the mismatch between two consecutive actions -- namely, the action \emph{Question Number} and \emph{Subtopic}. That is due to the order of these actions in the conversational flow: Occurrence of both actions in the dialogue is not strict and substantially depends on the previous system action. However, the analysis of the \emph{extended model} revealed that the introduction of additional context in the form of the previous action improved the performance of the system in this particular case by about $60\%$.

\textbf{Named Entity Recognition:} The failing cases include mismatches between the tags \emph{``chapter''} and \emph{``other''}, and the tags \emph{``question number''} and \emph{``other''}. This type of error arose due to the imperfectly labeled span of a multi-word named entity. In such cases, the first or last word in the named entity was excluded from the span and erroneously labeled with the tag \emph{``other''}.

\section{Related Work}
Individual components of a particular dialogue system could be implemented using a different kind of approach, starting with entirely \emph{rule- and template-based methods}, and going towards \emph{hybrid approaches} (using learnable components along with handcrafted units) and \emph{end-to-end trainable machine learning methods}.
\vspace{0.5mm}

{\bf Rule-based Approaches:}
Though many of the latest research approaches handle NLU and NLG units by using statistical NLP models \cite{bocklisch2017rasa,burtsev2018deeppavlov,honnibal2017spacy}, most of the  industrially deployed dialogue systems still use manual features or handcrafted rules for the state and action prediction, intent classification, and slot filling tasks \cite{chen2017survey,pydial}. The rule-based approach ensures robustness and stable performance that is crucial for industrial systems that interact with a large number of users simultaneously. However, it is highly expensive and time-consuming to deploy a real dialogue system built in this manner. The major disadvantage is that the usage of handcrafted systems is restricted to a specific domain, and possible domain adaptation requires extensive manual engineering.
\vspace{0.5mm}

{\bf End-to-End Learning Approaches:}
Due to the recent advance of \emph{end-to-end neural generative models} \cite{collobert2011natural}, many efforts have been made to build an end-to-end trainable architecture for dialogue systems. Rather than using the traditional pipeline, an end-to-end model is conceived as a single module \cite{chen2017survey}. Despite having better adaptability compared to any rule-based system and being easy to train, end-to-end approaches remain unattainable for commercial conversational agents operating on real-world data. A well and carefully constructed task-oriented dialogue system in a known domain using handcrafted rules and predefined responses, still outperforms the end-to-end systems due to its robustness \cite{wu2019global,glasmachers2017limits}.
\vspace{0.5mm}

{\bf Hybrid Approaches:}
Though end-to-end learning is an attractive solution for dialogue systems, current techniques are data-intensive and require large amounts of dialogues to learn simple actions. To overcome this difficulty, \citeauthor{williams2017hybrid} (\citeyear{williams2017hybrid}) introduce \emph{Hybrid Code Networks (HCNs)}, which is an ensemble of retrieval and trainable units. The authors report, that compared to existing end-to-end methods, their approach considerably reduces the amount of data required for training \cite{williams2017hybrid}. Hybrid models appear to replace the established rule- and template-based approaches which are currently utilized in an industrial setting.

\section{Conclusions}
\seclabel{concl}

In this work, we implemented a dialogue system for Intelligent Process Automation purposes that simultaneously solves two problems: First, it reduces repetitive and time-consuming activities and, therefore, allows workers of the e-learning platform to focus on solely mathematical and hence more cognitively demanding questions. Second, by interacting with users, it augments the resources with structured and labeled training data for further possible implementation of learnable dialogue components. The realization of such a system was connected with many challenges. Among others were missing structured data, ambiguous or erroneous user-generated text and the necessity to deal with already existing corporate tools and their design. The introduced model allowed us to accumulate structured and to some extent labeled data without any special efforts from the human (i.e., tutors) side (e.g., manual annotation of existing dialogues, change of the conversational structure). Once we collected structured dialogues, we were able to re-train specific components of the system with deep learning methods and achieved reasonable performance for all proposed tasks. 

We believe the obtained results are rather good, considering a relatively small amount of data we utilized to fine-tune the pre-trained model. We, therefore, conclude that both Next Action Prediction and Named Entity Recognition components could be employed to substitute or extend the existing rule-based modules. Rule-based units are restricted in their capabilities and could be hardly adaptable to novel patterns, whereas the trainable units generalize better, which we believe could reduce the number of erroneous predictions in case of unexpected dialogue behavior. Furthermore, to increase the overall robustness, both rule-based and trainable components could be used synchronously as a hybrid model: in  case when one system fails, the dialogue proceeds on the prediction obtained from the other model.

\section{Future Work}
The core of the rule-based model is a dialogue manager that determines the current state of the conversation and the possible next action. Rule-based systems are generally considered to be hardly adaptable to new domains; however, our dialogue manager proved to be flexible to slight modifications in a workflow. One of the possible directions of future work would be the investigation of the \emph{general adaptability} of the dialogue manager core to other scenarios and domains (e.g., different course).
Further investigation could be towards the multi-language modality for the re-implemented units. Since the \emph{OMB+} platform also supports English and Chinese, it would be interesting to examine whether the simple translation from target language (i.e., English, Chinese) to source language (i.e., German) would be sufficient to employ already-assembled dataset and pre-trained units.

\paragraph{Acknowledgments.}
We gratefully acknowledge the \emph{OMB+} team for the
collaboration and especially thank \emph{Ruedi Seiler} for his helpful feedback and technical support. We are indebted to the tutors for the
evaluation of the system, as well as to the anonymous reviewers for their valuable comments.

\bibliography{AAAI20}
\bibliographystyle{aaai}


\newpage
\clearpage
\appendix

\section{OMB+ Design}
\seclabel{ombfig}
\figref{fig:omb} presents an example of the OMB+ Online Learning Platform.

\begin{figure*}[ht!]
\centering
\includegraphics[scale=0.45]{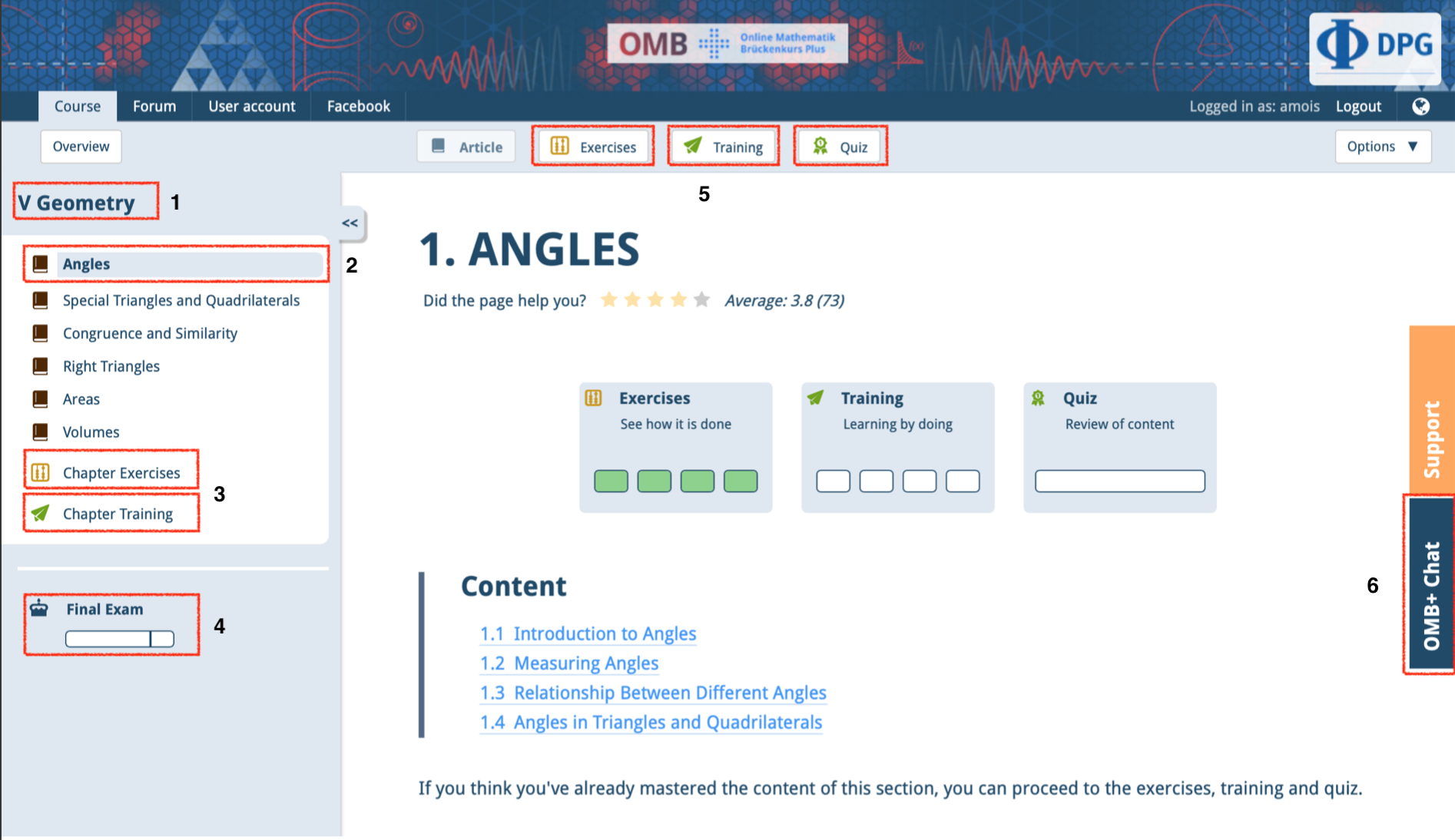}
\caption[\emph{OMB+} Online Learning Platform]{\emph{OMB+} Online Learning Platform, where \circled{1} is the \emph{Topic} (corresponds to a \emph{chapter level}), \circled{2} is a \emph{Sub-Topic} (corresponds to a \emph{section level}), \circled{3} is \emph{chapter level} examination mode, \circled{4} is the \emph{Final Examination Mode} (available only for \emph{chapter level}), \circled{5} are the \emph{Examination Modes}: Exercise, Training (available at \emph{section levels}) and \emph{Quiz} (available only at \emph{section level}), and \circled{6} is the \emph{OMB+ Chat}}
\figlabel{fig:omb}
\end{figure*}

\section{Mutually Exclusive Rules}
\seclabel{mers}

Assume a list of all theoretically possible dialogue states: $S$ = [topic, sub-topic, training, exercise, chapter level, section level, quiz, final examination, question number] and for each element $s_{n}$ in $S$ is true that: 

\begin{equation}
\label{states}
  s_{n}=\left\{
  \begin{array}{@{}ll@{}}
    1, & \text{if}\ \text{information is given}; \\
    0, & \text{otherwise}
  \end{array}\right.
\end{equation} 

This would give us all general (resp.~possible) dialogue states without reference to the design of the \emph{OMB+} platform. However, to make the dialogue states fully suitable for the \emph{OMB+}, from the general states, we take only those, which are \emph{valid}. To define the \emph{validness} of the state, we specify the following five Mutually Exclusive Rules (MER):

\begin{table}[h]
\centering
\begin{tabular}{@{}rr@{}}
\multicolumn{2}{r}{\bf $R_{1}$: Topic} \\ \hline\hline
$\neg$ T & T \\
\end{tabular}
\caption[Rule 1: Admissible configurations.]{Rule 1 -- Admissible \emph{topic} configurations.}
\tablabel{rule1}
\end{table}

Rule ($R_{1}$) in \tabref{rule1} denotes admissible configurations for \emph{topic} and means that we are either given a topic ($T$) or not ($\neg T$).

\begin{table}[h]
\centering
\begin{tabular}{@{}rrrr@{}}
\multicolumn{4}{r}{\bf $R_{2}$: Examination Mode} \\ \hline\hline
$\neg$ TR & $\neg$ E & $\neg$ Q & $\neg$ FE \\
TR & $\neg$ E & $\neg$ Q & $\neg$ FE \\
$\neg$ TR & E & $\neg$ Q & $\neg$ FE \\
$\neg$ TR & $\neg$ E & Q & $\neg$ FE \\
$\neg$ TR & $\neg$ E & $\neg$ Q & FE
\end{tabular}
\caption[Rule 2: Admissible configurations.]{Rule 2 -- Admissible \emph{examination mode} configurations.}
\tablabel{rule2}
\end{table}

Rule ($R_{2}$) in \tabref{rule2} denotes that either \emph{no} information on the examination mode is given, or examination mode is Training ($TR$) or Exercise ($E$) or Quiz ($Q$) or Final Examination ($FE$), but not more than one mode at the same time.

\begin{table}[h]
\centering
\begin{tabular}{@{}rr@{}}
\multicolumn{2}{r}{\bf $R_{3}$: Level} \\ \hline\hline
$\neg$ CL & $\neg$ SL  \\
CL & $\neg$ SL  \\
$\neg$ CL & SL
\end{tabular}
\caption[Rule 3: Admissible configurations.]{Rule 3 -- Admissible \emph{level} configurations.}
\tablabel{rule3}
\end{table}

Rule ($R_{3}$) in \tabref{rule3} indicates that either \emph{no} level information is provided, or the level corresponds to chapter level ($CL$) or to section level ($SL$), but not to both at the same time.

\begin{table}[h]
\centering
\begin{tabular}{@{}rrrr@{}}
\multicolumn{4}{r}{\bf $R_{4}$: Examination \& Level} \\ \hline\hline
$\neg$ TR & $\neg$ E & $\neg$ SL & $\neg$ CL \\
TR & $\neg$ E & SL & $\neg$ CL \\
TR & $\neg$ E & $\neg$ SL & CL \\
TR & $\neg$ E & $\neg$ SL & $\neg$ CL \\
$\neg$ TR & E & SL & $\neg$ CL \\
$\neg$ TR & E & $\neg$ SL & CL \\
$\neg$ TR & E & $\neg$ SL & $\neg$ CL
\end{tabular}
\caption[Rule 4: Admissible configurations.]{Rule 4 -- Admissible \emph{examination mode} (only for Training and Exercise) and corresponding \emph{level} configurations.}
\tablabel{rule4}
\end{table}

Rule ($R_{4}$) in \tabref{rule4} means that Training ($TR$) and Exercise ($E$) examination modes can either belong to chapter level ($CL$) or to
section level ($SL$), but not to both at the same time.

\begin{table}[h]
\centering
\begin{tabular}{@{}rr@{}}
\multicolumn{2}{r}{\bf $R_{5}$: Topic \& Sub-Topic} \\ \hline\hline
$\neg$ T & $\neg$ ST  \\
T & $\neg$ ST  \\
$\neg$ T & ST \\
T & ST
\end{tabular}
\caption[Rule 5: Admissible configurations.]{Rule 5 -- Admissible \emph{topic} and corresponding \emph{sub-topic} configurations.}
\tablabel{rule5}
\end{table}

Rule ($R_{5}$) in \tabref{rule5} symbolizes that we could be either given only a topic ($T$) or the combination of topic and sub-topic ($ST$) at the same time, or only sub-topic, or no information on this point at all.

We then define a \emph{valid dialogue state}, as a dialogue state that meets all requirements of the abovementioned rules:

\begin{align}
    \forall s(\text{State}(s) \wedge R_{1-5}(s)) \to \text{Valid}(s)
\end{align}

After we get the valid states for our dialogues, we want to make a \emph{mapping} from each valid dialogue state to the \emph{next possible systems action}. For that, we first define five \emph{transition rules} \footnote{The order of transition rules is important.}:

\begin{align}
    T_{1}:= \neg \text{T},
\end{align}

means that no topic ($T$) is found in the ID (i.e., could not be extracted from user input).

\begin{align}
    T_{2}:= \neg \text{EM} ,
\end{align}

indicates that no examination mode ($EM$) is found in the ID.

\begin{align}
    T_{3}:= \text{EM, where EM} \in [TR, E],
\end{align}

denotes that the extracted examination mode ($EM$) is either Training ($TR$) or Exercise ($E$).

\begin{align}
\label{states}
  T_{4}:=\left\{
  \begin{array}{@{}rrrr@{}}
    T & \neg ST & TR & SL, \\
    T & \neg ST & E & SL, \\
    T & \neg ST & Q
  \end{array}\right.
\end{align}

means that \emph{no} sub-topic ($ST$) is provided by a user, but ID either already contains the combination of topic ($T$), training ($TR$) and section level ($SL$), or the combination of topic, exercise ($E$) and section level, or the combination of topic and quiz ($Q$). 

\begin{align}
\label{transition4}
T_{5}:= \neg \text{QNR} ,
\end{align}

indicates that \emph{no} question number ($QNR$) was provided by a student (or could not be successfully extracted).

Finally, we assumed the list of possible next actions for the system:
\begin{align}
\label{states}
  A=\left\{
  \begin{array}{@{}l@{}}
    \text{Ask for Topic,} \\
    \text{Ask for Examination Mode,} \\
    \text{Ask for Level,} \\
    \text{Ask for Sub-Topic,} \\
    \text{Ask for Question Nr.}
  \end{array}\right.
\end{align}

Following the transition rules, we mapped each valid dialogue state to the possible next action $a_{m}$ in $A$:  

\begin{align}
\exists a \forall s (\text{Valid}(s) \wedge T_{1} (s)) \to a(s, \text{T}), 
\end{align}

in the case where we do \emph{not} have any topic provided, the next action is to \emph{ask for the topic} ($T$).

\begin{align}
\exists a \forall s (\text{Valid}(s) \wedge T_{2} (s)) \to a(s, \text{EM}), 
\end{align}

if \emph{no} examination mode is provided by a user (or it could not be successfully extracted from the user query), the next action is defined as \emph{ask for examination mode} ($EM$).

\begin{align}
\exists a \forall s (\text{Valid}(s) \wedge T_{3} (s)) \to a(s, \text{L}), 
\end{align}

in case where we know the examination mode $EM$ $\in$ [Training, Exercise], we have to ask about the level (i.e., training at chapter level or training at section level), thus the next action is \emph{ask for level} ($L$).

\begin{align}
\exists a \forall s (\text{Valid}(s) \wedge T_{4} (s)) \to a(s, \text{ST}), 
\end{align}

if \emph{no} sub-topic is provided, but the examination mode $EM$ $\in$ [Training, Exercise] at \emph{section level}, the next action is defined as \emph{ask for sub-topic} ($ST$).

\begin{align}
\exists a \forall s (\text{Valid}(s) \wedge T_{5} (s)) \to a(s, \text{QNR}), 
\end{align}

if \emph{no} question number is provided by a user, then the next action is \emph{ask for question number} ($QNR$).

\section{Completeness of ID: Example Cases }
\seclabel{ap:a_cases}

Below are examples of two final cases (out of six), where ID is considered to be complete:

\begin{table}[h] 
\begin{align}
\label{states}
  Case_1=\left\{
  \begin{array}{@{}l@{}}
    \text{intent = Math,} \\
    \text{topic $\ne$ None,} \\
    \text{exam mode = Final Examination,} \\
    \text{level = None,} \\
    \text{question nr. $\ne$ None}
  \end{array}\right.
\end{align}
\caption{Case 1 -- Any topic, examination mode is final examination, examination level does not matter, any question number.}
\end{table}

\begin{table}[h] 
\begin{align}
\label{states}
  Case_2=\left\{
  \begin{array}{@{}l@{}}
    \text{intent = Math,} \\
    \text{topic $\ne$ None,} \\
    \text{sub-topic $\ne$ None,} \\
    \text{exam mode $\in$ [Training, Exercise],} \\
    \text{level = Section,} \\
    \text{question nr. $\ne$ None}
  \end{array}\right.
\end{align}
\caption{Case 2 -- Any topic, any related sub-topic, examination mode is either training or exercise, examination level is section, any question number.}
\end{table}

\section{Interaction Showcases}
\seclabel{ap:a_appendix}

Below are five sample dialogues with variable flows.\footnote{Note, that the dialogues were translated from German to English for the readers convenience.}

\begin{table}[ht!]
\small
\fbox{\begin{minipage}{.5\textwidth}
\textbf{Student:} Hi, I have a question regarding the \underline{certificate issue}.

\vspace{0.4em}
\textbf{Next-Action:} [Organisational Question] \\
\textbf{Bot:} It seems to be an organisational question! Please, summarize your question in a short text so that I can forward it to my human colleague.

\vspace{0.4em}
\textbf{Student:} How should I proceed to issue my certificate?

\vspace{0.4em}
\textbf{Next-Action:} [Human Handover] \\
\textbf{Bot:} Thank you, our human colleagues will get back to you soon!
\end{minipage}}
\caption{Showcase 1 -- Organisational Question. Underlined are the key-words which point on the organisational intent.}
\label{d1}
\end{table}

\begin{table}[ht!]
\small
\fbox{\begin{minipage}{.5\textwidth}
\textbf{Student:} Hi, I have difficulties in \underline{Chapter I, Elementary Calculus}, by solving \underline{Exercise 1 a}

\vspace{0.4em}
\textbf{Next-Action:} [Ask for a Level: Chapter or Section]\\
\textbf{Bot:} Are you working on a task at the chapter-level (e.g. any training in the Chapter ``Geometry'') or at the section-level (e.g. Section ``Angle'' in the Chapter ``Geometry'')? Please answer with CHAP for chapter and SEC for section.

\vspace{0.4em}
\textbf {Student:} Chapter

\vspace{0.4em}
\textbf{Next-Action:} [Final Request] \\
\textbf {Bot:} You provided following information: \\
a) Chapter: I  Elementary Calculus,\\
b) Exercise Mode: Exercise,\\
c) Question Number: 1a,\\
d) Level: Chapter.\\
Did I understand you correctly? Please, answer Yes or No.

\vspace{0.4em}
\textbf{Student:} yep

\vspace{0.2em}
\textbf{Next-Action:} [Exact Question] \\
\textbf{Bot:} Please, summarize your question in a short text so that I can forward it to my human colleague.

\vspace{0.4em}
\textbf{Student:} I do not understand how to solve the exercise...

\vspace{0.4em}
\textbf{Next-Action:} [Human Handover] \\
\textbf{Bot:} Thank you, our human colleagues will get back to you soon!
\end{minipage}}
\caption{Showcase 2 -- Short flow, most of the information was extracted from the initial user input. Underlined are the extracted entities.}
\label{d2}
\end{table}

\begin{table}[ht!]
\small
\fbox{\begin{minipage}{.5\textwidth}
\textbf{Student:} Hi

\vspace{0.4em}
\textbf{Next-Action:} [UNK] \\
\textbf{Bot:} Do you have a mathematical question (MATH), a contextual question (TEXT) or an organizational question (ORG)?

\vspace{0.4em}
\textbf {Student:} ma

\vspace{0.4em}
\textbf{Next-Action:} [UNK] \\
\textbf{Bot:} Do you have a mathematical question (MATH), a contextual question (TEXT) or an organizational question (ORG)?

\vspace{0.4em}
\textbf {Student:} math

\vspace{0.4em}
\textbf{Next-Action:} [Ask for Topic] \\
\textbf{Bot:} Which Chapter you are working on right now?

\vspace{0.4em}
\textbf{Student:} Linear Systems of Equations

\vspace{0.4em}
\textbf{Next-Action:} [Ask for Examination Mode] \\
\textbf{Bot:} Are you working on an exercise, training, quiz or final examination?

\vspace{0.4em}
\textbf{Student:} \underline{human}

\vspace{0.4em}
\textbf{Next-Action:} [Human Hand-Over] \\
\textbf{Bot:} Thank you, our human colleagues will get back to you soon!
\end{minipage}}
\caption{Showcase 3 -- Manual \emph{intent classification}, \emph{Fallback} and \emph{Human Request} policies (underlined).}
\label{d3}
\end{table}

\vspace{4cm}
\begin{table}[ht!]
\small
\fbox{\begin{minipage}{.5\textwidth}

\textbf{Student:} Hi-Hi, could someone please explain me the \underline{rule} 7.2 (vertical stretching and compression of the graph)? Thank you in advance!

\vspace{0.4em}
\textbf{Next-Action:} [Contextual Question] \\
\textbf{Bot:} It seems to be a contextual question! Please, summarize your question in a short text so that I can forward it to my human colleague.

\vspace{0.4em}
\textbf{Student:} I'm having trouble understanding Rule 7.2. What exactly is the vertical extension of the graph?

\vspace{0.4em}
\textbf{Next-Action:} [Human Hand-Over] \\
\textbf{Bot:} Thank you, our human colleagues will get back to you soon!
\end{minipage}}
\caption{Showcase 4 -- Contextual Question. Underlined are the key-words which point on the contextual intent.}
\label{d4}
\end{table}
\newpage 

\begin{table}[ht!]
\small
\fbox{\begin{minipage}{.5\textwidth}

\textbf{Student:} Hi, I do not understand the \underline{Training 1 (a)} in \underline{Chapter 1}

\vspace{0.4em}
\textbf{Next-Action:} [Ask for Level] \\
\textbf{Bot:} Are you working on a task at the chapter-level (e.g. any training in the Chapter ``Geometry'') or at the section-level (e.g. Section ``Angle'' in the Chapter ``Geometry'')? Please answer with CHAP for chapter and SEC for section.

\vspace{0.4em}
\textbf {Student:} I think it is \underline{section}

\vspace{0.4em}
\textbf{Next-Action:} [Ask for Sub-Topic] \\
\textbf{Bot:} Which Section you are working on right now then?

\vspace{0.4em}
\textbf{Student:} I am working on \underline{roots and powers}

\vspace{0.4em}
\textbf{Next-Action:} [Final Request] \\
\textbf {Bot:} You provided following information: \\
a) Chapter: I Elementary Calculus,\\
b) Section: Calculate with Roots, Powers,\\
c) Examination Mode: Training,\\ 
d) Question Number: 1,\\
e) Level: Section.\\
Did I understand you correctly? Please, answer Yes or No.

\vspace{0.4em}
\textbf{Student:} nope

\vspace{0.4em}
\textbf{Next-Action:} [Verify Request] \\
\textbf{Bot:} Which of the following items are misrecognized? Please, specify: a, b, c, d, e?

\vspace{0.4em}
\textbf{Student:} d

\vspace{0.4em}
\textbf{State-Transition:} [Correct Request] \\
\textbf{Bot:} Please, provide the correct information for d) :

\vspace{0.4em}
\textbf{Student:} 1 (a)

\vspace{0.4em}
\textbf{State-Transition:} [Final Request]\\
\textbf {Bot:} You provided following information: \\
a) Chapter: I Elementary Calculus,\\
b) Section: Calculate with Roots, Powers,\\
c) Examination Mode: Training,\\ 
d) Question Number: 1 (a),\\
e) Level: Section.\\
Did I understand you correctly? Please, answer Yes or No.

\vspace{0.4em}
\textbf{Student:} yes 

\vspace{0.2em}
\textbf{Next-Action:} [Exact Question] \\
\textbf{Bot:} Please, summarize your question in a short text so that I can forward it to my human colleague.

\vspace{0.4em}
\textbf{Student:} I do not understand how to solve the exercise...

\vspace{0.4em}
\textbf{State-Transition:} [Human Handover] \\
\textbf{Bot:} Thank you, our human colleagues will get back to you soon!
\end{minipage}}
\caption{Showcase 5 -- Long Flow. Correction of entries. Underlined are the extracted entities.}
\label{d5}
\end{table}

\begin{figure*}[!hb]
\centering
\includegraphics[scale=0.2]{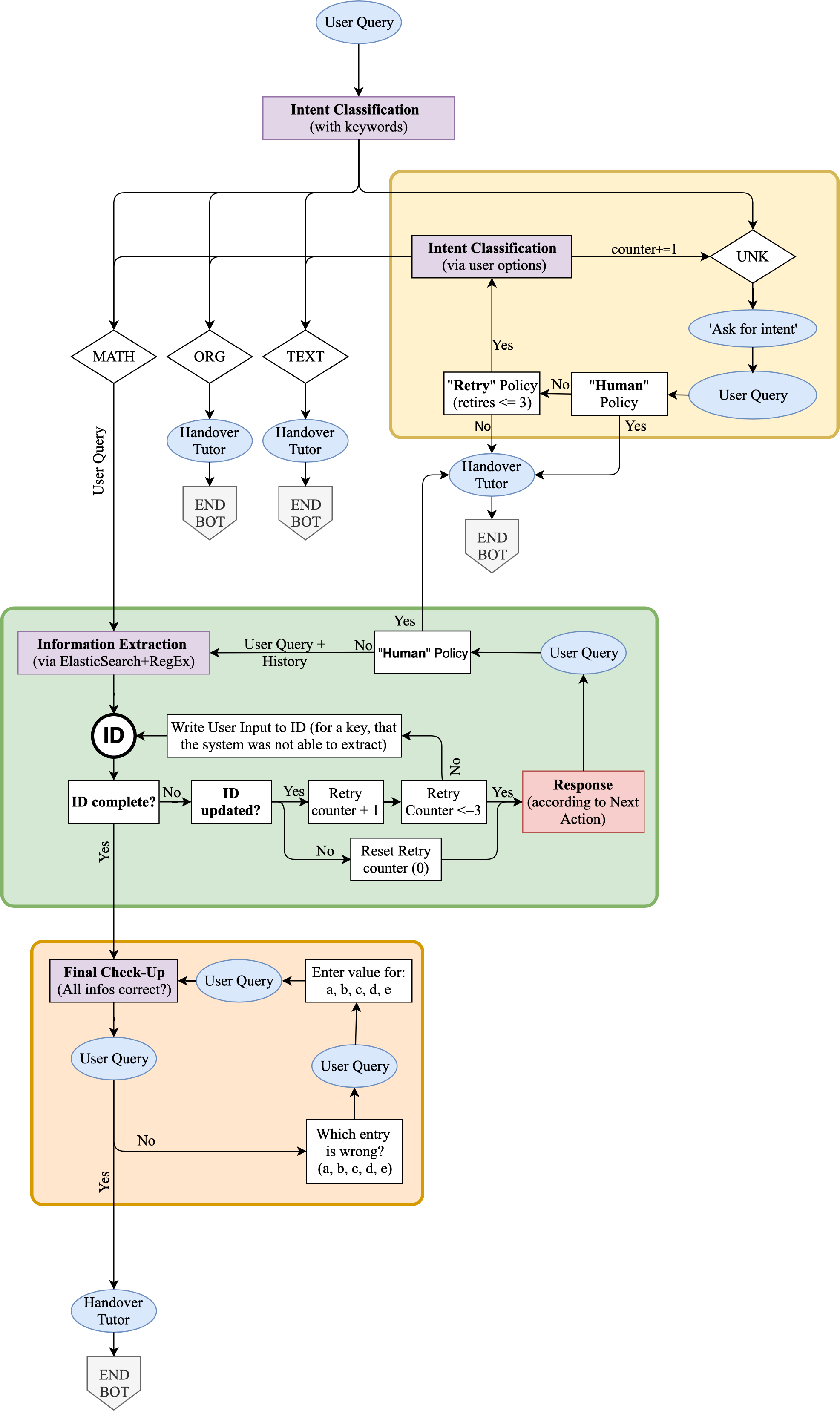}
\caption[Ruled-Based System. Dialogue Flow]{Dialogue Flow. Abbreviations: UNK - unknown; ID - information dictionary; RegEx - regular expressions.}
\figlabel{fig:df}
\end{figure*}

\end{document}